\pdfoutput=1

\documentclass[11pt]{article}

\usepackage{ACL2023}
\usepackage{graphicx}
\usepackage{amssymb}

\usepackage{times}
\usepackage{latexsym}

\usepackage[T1]{fontenc}

\usepackage[utf8]{inputenc}

\usepackage{microtype}

\usepackage{inconsolata}

\title{Intelligence Analysis of Language Models}

  \author{Liane Galanti \\
  \texttt{lianegalanti@mail.tau.ac.il} \\\And
Ethan Baron \\
  \texttt{ethanbarbar@gmail.com} \\}

\begin{document}
\maketitle

\begin{abstract}
In this project, we test the effectiveness of Large Language Models (LLMs) on the Abstraction and Reasoning Corpus (ARC) \cite{Chollet2019OnTM} dataset. This dataset serves as a representative benchmark for testing abstract reasoning abilities, requiring a fundamental understanding of key concepts such as object identification, basic counting, and elementary geometric principles. Tasks from this dataset are converted into a prompt-based format for evaluation.  Initially, we assess the models' potential through a Zero-shot approach. Subsequently, we investigate the application of the Chain-of-Thought (CoT) \cite{COT} technique, aiming to determine its role in improving model performance. Our results suggest that, despite the high expectations placed on contemporary LLMs, these models still struggle in non-linguistic domains, even when dealing with simpler subsets of the ARC dataset. 
Our study is the first to concentrate on the capabilities of open-source models in this context. The code, dataset, and prompts supporting this project's findings can be found in our GitHub repository, accessible at: \\
\href{https://github.com/Lianga2000/LLMsOnARC}{LLMsOnARC}.

\end{abstract}


\section{Introduction and Related Work}

Evaluating artificial intelligence (AI) system intelligence through abstract and visual reasoning challenges has been a longstanding practice in the field of deep learning. An early instance is the Analogy program by Evans \cite{Evans1964APF}, which solved geometric analogy tasks using DSL. Numerous programs and benchmarks have been proposed over the years, with the ARCathon \cite{Chollet2019OnTM} being the latest and most prominent example.

\paragraph{ARC dataset and ARCathon}
The Abstraction and Reasoning Corpus (ARC) created by François Chollet \cite{Chollet2019OnTM} is a benchmark test that measures AI's progress towards human-level abstraction and reasoning abilities. It resembles in format to Raven’s Progressive Matrices \cite{raven}, a traditional IQ test format that has been in use since the 1930s. ARCathon, is a global AI competition. This highly challenging event welcomes participants to tackle the ARC challenge unsolvable for the most advanced deep learning models currently available. 
\\Every intelligence test inherently relies on some prior knowledge. ARC addresses this by specifying the priors it is based on, ensuring that it doesn't depend on information outside these defined priors, such as language or other acquired knowledge. The ARC's priors are made to be very similar to what is called Core Knowledge priors. This is done for making a fair comparison between the intelligence of humans and that made by machines (artificial intelligence).
The priors in the ARC framework:
\begin{itemize}
    \item Objectness priors: Recognition of objects based on color and shape, their persistence despite changes (see figure \ref{fig:denoising_task}), and interaction through physical contact (see figure \ref{fig:move_red}).

\begin{figure}
  \centering
  \includegraphics[width=0.5\textwidth]{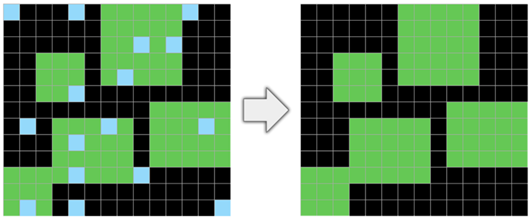}
  \caption{A denoising task from the ARC dataset.}
   \label{fig:denoising_task}
\end{figure}
\begin{figure}
  \centering
  \includegraphics[width=0.5\textwidth]{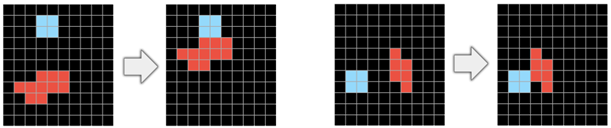}
  \caption{A task from ARC where you need to move the red object towards the blue object until they contact.}
   \label{fig:move_red}
\end{figure}
    \item Goal-Directedness prior: Interpreting tasks as processes with intentional start and end states (see figure \ref{fig:moving_obj}), despite the absence of a time concept.

\begin{figure}
  \centering
  \includegraphics[width=0.5\textwidth]{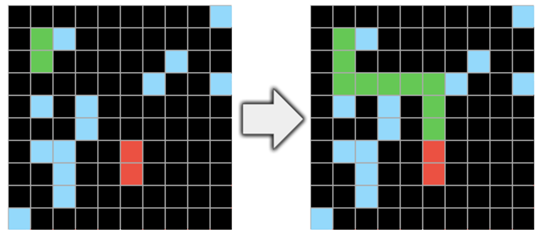}
  \caption{A task from ARC where you need connect starting point (green object), to the ending point (the red object) and you can turn only when touching a blue object.}
   \label{fig:moving_obj}
\end{figure}

    \item Numbers and Counting priors: Involving tasks with counting, sorting, and basic arithmetic (see figure \ref{fig:counting_obj}).

\begin{figure}
  \centering
  \includegraphics[width=0.5\textwidth]{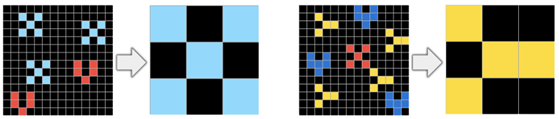}
  \caption{A task from ARC where you need to count unique objects and select the object that appears the most times.}
   \label{fig:counting_obj}
\end{figure}

    \item Basic Geometry and Topology priors: Involves elementary geometry and topology concepts like lines, rectangles, symmetries (see figure \ref{fig:symmetry}), rotations, scaling, distortions, containing/being contained, drawing lines, orthogonal projections, and object repetition.

\begin{figure}
  \centering
  \includegraphics[width=0.5\textwidth]{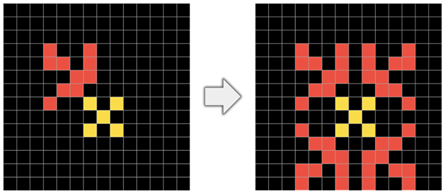}
  \caption{A task from ARC where you need to draw symmetrically a shape around a marker.}
   \label{fig:symmetry}
\end{figure}

\end{itemize}


\paragraph{Alternate Approaches to ARCathon Besides Deep Learning} Since ARC's inception in 2019 \cite{Chollet2019OnTM}, diverse methods have emerged. For instance, the winning solution in the Kaggle challenge \cite{kaggle0} employed a handcrafted DSL and DAG-based search. Similar strategies were followed by other top Kaggle entries. In another approach, \cite{grammar-arc} used the Grammatical Evolution algorithm within their DSL, while \cite{dreamcoderarc} leveraged the DreamCoder program synthesis system \cite{dreamcoder} to generate solutions through neural-guided synthesis. Recently, ARGA \cite{arga} introduced an object-centric framework using graph representation and tree search for a DSL based on abstracted graph space.

\paragraph{Deep Learning Approaches} 
Deep learning methods have been investigated to tackle the ARC challenge. The Neural Abstract Reasoner, a deep learning method, managed to solve some ARC tasks successfully \cite{Kolev2020}. Another study \cite{Assouel2020} introduced a technique aiming to create new tasks for improved generalization. As LLMs alleged reasoning capabilities increased dramatically over recent years, some attempted to use them for solving the ARC challenge. An interesting attempt is by \cite{conceptarc} which created an interesting similar dataset named ConceptArc. They employed ARC solvers and GPT-4 \cite{GPT4} to tackle tasks from the ConceptARC dataset, comparing these solutions with those produced by humans. To enhance the performance of LLMs, various prompting techniques have been developed. One example is the Chain-of-Thought (CoT) approach \cite{COT}, which involves providing LLMs with intermediate reasoning steps. In our study, we evaluate the performance of LLMs directly on ARC tasks (without simplifying the tasks), employing Zero-shot and CoT as prompting methods. To our knowledge, this is the first study focusing exclusively on open-source models for such an evaluation. The LLMs tested include LLaMA \cite{LLAMA}, Phind (fine-tuned version of CodeLlama-34B), and Mixtral \cite{MIXTRAL}.

\paragraph{Simpler ARC}
To better comprehend the complexity of ARC, datasets resembling ARC but with simplified challenges have been developed. Mini-ARC \cite{MiniARC} offers a condensed version with $5x5$ matrices. The ConceptARC \cite{conceptarc} dataset presents a collection of manually designed tasks. Similarly, Sort-of-ARC \cite{Assouel2020} adopts the ARC input format but features less complex challenges, comprising $20x20$ images containing three distinct $3x3$ objects. Another dataset includes a subset of 50 ARC tasks solvable by the ARC solver ARGA \cite{arga}, following the approach outlined by \cite{GPTonARC}. We specifically focused on assessing the performance of various LLMs on these selected tasks.

\section{Method}
A straightforward approach to addressing ARC-like tasks with LLMs is through textual encoding. This technique involves transforming the 2D input-output images into a textual format. Once converted, this text becomes part of the LLM's prompt. The LLM then processes this prompt and generates a solution that hopefully corresponds to the required output.

\subsection{Dataset}
A subset of 50 ARC tasks that are solvable by ARC solver ARGA \cite{arga}. This method follows what was first done in \cite{GPTonARC}. This subset was curated specifically because it is shorter and "simpler", as these tasks can be solved using a search-based approach. \\Each task consists of a small number of demonstration examples and a small number of test examples (generally 1). Each example consists of an input matrix and an output matrix. Each matrix is of dimension $nxm$ ($n,m$ are the number of rows and columns of the matrix accordingly) and each element in it is a color (e.g., see figure \ref{fig:counting_obj}). There 10 different colors. A matrix can have any height or width between $1-30$.

\subsection{Textual encoding}
Given an ARC task, i.e., 2D matrix, we converted it into text by encoding each element color numerically from "0" to "9", representing ten colors. In our encoding, numbers are space-separated, and matrix rows are marked with "\textbackslash n". Alternatively, colors could be encoded with letters "a" to "j" or by their names, like "blue."

\subsection{Prompting techniques}
\paragraph{Zero-shot:}
For this method, the models were given the tasks as prompts and were asked to produce outputs without prior training. This approach evaluates the innate ability of LLMs to tackle new, reasoning-based challenges. It provides valuable insights into their adaptability and generalization capabilities. Such an evaluation is important in understanding how well these models can perform in real-world scenarios where they encounter unexpected problems.  


\paragraph{Chain-of-Thought (CoT):} 
For this method, the models were given the tasks and a step-by-step reasoning process tailored for a specific fixed task from the dataset. These were provided as prompts, and the models were then asked to generate the corresponding output. It's important to note that the models were not tested on this fixed task. 
\\ARC tasks typically require advanced, multi-step reasoning. The CoT \cite{COT} method is designed to guide the models through the necessary steps to solve such complex tasks. By training with CoT, models are expected to better understand the multi-step problem-solving, a critical skill for successfully handling ARC dataset.

\section{Results}
\begin{table}[h]
\centering
\begin{tabular}{|l|c|c|}
\hline
\textbf{Model} & \textbf{Zero-shot} & \textbf{CoT} \\
\hline
Code Llama 7-b & 2 & 1 \\
\hline
Code Llama 13-b & 1 & 2 \\
\hline
Phind & 2 & 2 \\
\hline
Mixtral 8x7b & 2 & 2 \\
\hline
\end{tabular}
\caption{Performance comparison. Each column represents a different prompting method, and each row is assigned to a specific model. 
The values correspond to the number of tasks, out of 50, solved by each method. These are the best results obtained from the three tested
encoding techniques.}
\label{table:model_performance}
\end{table}

\begin{table}[h]
\centering
\begin{tabular}{|l|c|c|c|c|}
\hline
\textbf{Model \textbackslash task \#} & \textbf{3} & \textbf{14} & \textbf{17} & \textbf{47} \\
\hline
Code Llama 7-b & \checkmark & \checkmark & $\times$ & $\times$ \\
Code Llama 7-b using CoT  & \checkmark & $\times$ & $\times$ & $\times$ \\
\hline
Code Llama 13-b  & \checkmark & $\times$ & $\times$ & $\times$ \\
Code Llama 13-b using CoT & \checkmark & \checkmark & $\times$ & $\times$ \\
\hline
Phind & \checkmark & \checkmark & $\times$ & $\times$ \\
Phind using CoT & $\times$ & \checkmark & $\times$ & \checkmark \\
\hline
Mixtral & \checkmark & \checkmark & $\times$ & $\times$ \\
Mixtral using CoT  & \checkmark & $\times$ & \checkmark & $\times$ \\
\hline
\end{tabular}
\caption{Task Success by Model and Method. This table compares the performance of each model using either Zero-shot or Chain-of-Thought (CoT) prompting methods on specific ARC tasks, numbered 3, 14, 17, and 47. These tasks are highlighted as they are the only ones in the dataset where models achieved success; all other tasks resulted in failures. A checkmark (\checkmark) indicates successful completion of the task, while a cross ($\times$) denotes failure.}
\label{table:model_success}
\end{table}

\begin{figure}
  \centering
  \caption*{Left Side \hfill Right Side  \hfill \hfill }
  \includegraphics[width=0.5\textwidth]{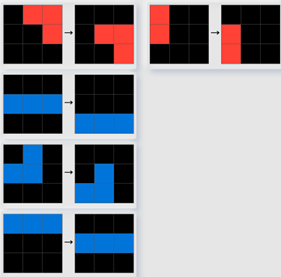}
  \caption{Task \#3. On the left are the demonstration examples, and on the right is the test example, including both the task and its expected output. This task was successfully solved by all models, with the exception of Phind when using the CoT method.}
  \label{fig:easy_task}
\end{figure}

\begin{figure}
  \centering
  \caption*{Left Side \hfill Right Side  \hfill \hfill }
  \includegraphics[width=0.5\textwidth]{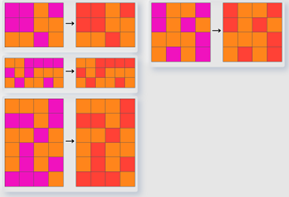}
  \caption{Task \#17. Left side is the demonstration examples and right side is the test example (the task and its expected output). This task was only solved by Mixtral using CoT. }
   \label{fig:task17}
\end{figure}
\begin{table}
  \centering
  \begin{tabular}{|l|}
  \hline
    By looking at the given input-output pairs, I can \\ infer that the underlying abstract transformation \\ is as follows: \\
    \hline
    - For each row in the input: \\
    \hline
    - If the row is all zeros, leave it unchanged \\in the output. \\  
    \hline
    - Otherwise, shift the non-zero values in the row \\to the right by one position, and fill the leftmost \\ position with a zero.
    \\
    \hline
    Applying this transformation to the given input:
    \\
    \hline
    input:
    \\
    \hline$[[200][200][000]]$
    \\
    \hline output:
    \\
    \hline$[[000][200][200]]$
    \\
    \hline
    Explanation:
    \\
    \hline
    - The first row $[200]$ is not all zeros, so we \\ shift its non-zero value to the right by one \\ position, resulting in $[020]$.
    \\
    \hline- The second row $[200]$ is not all zeros, so we \\ shift its non-zero value to the right by \\ one position, resulting in $[020]$.
    \\
    \hline- The third row $[000]$ is all zeros, so we leave it \\ unchanged.
    \\
    \hline
    So the output is:
    \\
    \hline
    $[[000][200][200]]]$
    \\
    \hline
  \end{tabular}
  \caption{The reasoning provided by the model Mixtral using CoT on the task \#3 shown in figure \ref{fig:easy_task}. As we can see the reasoning is wrong even though the output is correct.}
  \label{table:badCoT}
\end{table}

\begin{table}
  \centering
  \begin{tabular}{|l|}
  \hline

    By looking at the given input-output pairs, I can 
    \\infer that the input and output are two-dimensional
    \\grids of numbers, the transformation applied is as follows: \\
    \hline
    - Each 6 in the input grid is transformed to 2\\
    in the output grid. \\
    \hline
    -- Each 7 in the input grid is transformed to 
    \\the same number in the output grid output. \\  
    \hline
    - Each 67 pair in the input grid is\\
    transformed to 72 in the output grid.
    \\
    \hline
    - Each 76 pair in the input grid is 
    \\transformed to 27 in the output grid.
    \\
    \hline
    Thus, for the test input:
    \\
    $[[6776][6767][7776]$
    \\
    \hline output matrix is:
    \\
    \hline$[[2772][2727][7772][7272]]]$
    \\
    \hline
  \end{tabular}
  \caption{The reasoning provided by the model Mixtral using CoT on the task \#17, shown in figure \ref{fig:task17}, indicates that the reasoning is only partly correct, even though the output is accurate.}
  \label{table:badCoT2}
\end{table}

The results indicate that all the models we have tested, showed limited effectiveness in solving the ARC tasks. The maximum number of tasks successfully solved by any model was 2 out of a total of 50 (see table \ref{table:model_performance}), suggesting a notable challenge for LLMs in this particular problem domain.
\\
The comparison of Zero-shot and CoT approaches in terms of performance (see table \ref{table:model_success}) suggests that the effectiveness of CoT is somewhat questionable. It is observed that models using CoT don't consistently reason correctly towards the right solution, even when they eventually provide the correct answer e.g., see tables \ref{table:badCoT} and \ref{table:badCoT2}. Moreover, Code Llama 7-b performed better under Zero-shot than CoT, while Code Llama 13-b showed the opposite trend.
\\
Analyzing the success across specific tasks numbered 3, 14, 17, 47 (see table \ref{table:model_success}), it's observed that certain tasks (e.g., task \#3, see figure \ref{fig:easy_task}) were consistently solved across different models and prompting methods, indicating these tasks might be easier to solve or more aligned with the models' capabilities. In contrast, tasks like \#17 and \#47 were rarely or never solved, pointing to their higher complexity or misalignment with the models' reasoning patterns.

\subsection{Experimental Details}

\paragraph{Models}
In our study, we evaluated several models: LLaMA \cite{LLAMA}, a META-owned LLM known for its robust performance across a range of benchmarks; We also assessed Phind which is an adaptation of CodeLlama-34B, fine-tuned on a dataset exclusive to Hugging Face; and the recently released Mixtral \cite{MIXTRAL}, famous for its speed, success across a range of benchmarks and versatility in language.

\paragraph{Compute}
Each task was executed three times. A task was considered successfully completed by a model if it achieved success in at least one of these attempts. Due to constraints in GPU resources, all runs were conducted using 8-bit quantization on two Quadro RTX 8000 GPUs, each with 50GB of memory.

\section{Discussion} 
In our study, we specifically investigated the capability of Large Language Models (LLMs) in tasks that require abstract reasoning. This systematic analysis is unique in its focus on open-source LLMs, utilizing Zero-shot and Chain-of-Thought (CoT) as prompting techniques. The results clearly show that the models underperform on the dataset, achieving a maximum success rate of only 2 out of 50 tasks. These findings align with recent similar research, such as the study described in \cite{GPTonARC}, which exclusively evaluated GPT \cite{GPT3, GPT4} models on the 50 tasks subset. In one of their configurations, GPT-3.5 \cite{GPT3} achieved success in only 3 out of 50 tasks using Zero-shot and 2 using CoT. For GPT-4 \cite{GPT4}, the success rates were slightly higher, with 5 tasks solved using Zero-shot and 9 with CoT.
An interesting observation of our study, also observed in \cite{GPTonARC}, is that employing CoT sometimes resulted in poorer outcomes (as shown in tables \ref{table:model_performance} and \ref{table:model_success}). In both our study and in \cite{GPTonARC}, it was evident that even when models correctly solved tasks using CoT, their reasoning processes were not necessarily sound or logical (see tables \ref{table:badCoT}, \ref{table:badCoT2}). This suggests that while CoT can lead to correct answers, it does not always ensure a coherent or accurate reasoning, highlighting a significant limitation in its effectiveness. These outcomes suggest that LLMs generally struggle with abstract reasoning tasks, as demonstrated by their limited success on the ARC dataset.

It is important to underscore that the annual ARCathon competition, dedicated to advancing models capable of handling the ARC dataset, has not seen any significantly improved outcomes since 2020 with a success rate of $\sim30\%$ on the dataset (for more detailes, see \href{https://innovators-guide.ch/2023/02/arcathon-global-ai-competition-davos/}{ARCathon}). Notably, the solution that achieved this did not employ any deep learning models. When deep learning methods have been applied to this challenge, they have shown only minimal, if any, enhancement in performance. For instance, in a study conducted by \cite{GPTonARC}, where GPT was tested on ARC, the success rate was far below 30\%. This lack of advancement highlights a critical aspect: despite progress in deep learning and particularly in language models, LLMs remain challenged in abstract reasoning. Our analysis contributes to the growing body of evidence suggesting that current LLMs are a considerable distance from achieving Artificial General Intelligence (AGI). The consistent inability of LLMs to excel in abstract reasoning, even when leveraging techniques like CoT, highlights an important limitation in their current design and functioning.

While our study represents a step forward in understanding how Large Language Models (LLMs) handle reasoning, it is important to recognize its limitations. We relied on only two prompting techniques, and it is possible that alternative techniques could yield improved results. Furthermore, our focus was limited to open-source LLMs, which may not encompass the full spectrum of capabilities present in more advanced models. This limitation can be seen by the better results achieved with GPT \cite{GPT3, GPT4}, as demonstrated in \cite{GPTonARC}. 
Another limitation was the focus on only the 50 easiest tasks in ARC, standard approach took place in \cite{GPTonARC}. This approach underlines the fact that LLMs still have a considerable way to go in solving complex tasks that necessitate abstract and visual reasoning skills.

Future work could expand on our findings by exploring a variety of prompting techniques, such as Tree-of-Thoughts (ToT) \cite{ToT}, an extension of CoT \cite{COT}. 
Additionally, within the CoT method, a broader selection of examples from the dataset could be included in the prompts. For instance, instead of using just one example task as in our study, including multiple example tasks in the prompt that correspond to each of ARC's core knowledge priors — each with a tailored step-by-step reasoning process — could potentially improve the models' ability to successfully tackle a wider range of tasks from the dataset. 
Furthermore, the use of more advanced models, perhaps fine-tuned for abstract reasoning, could be beneficial.
There is also a potential to evaluate LLMs against simplified versions of the ARC dataset, including those proposed by \cite{MiniARC, conceptarc, Assouel2020}. Additionally, comparative studies involving human performance on similar tasks, as attempted in \cite{conceptarc}, could provide insightful benchmarks for assessing the capabilities of LLMs.


To conclude, our study sheds light on the current state of LLMs in handling complex tasks like those in the ARC dataset. Our findings indicate that while LLMs have made remarkable steps in many areas, they are not yet equipped to effectively tackle complex tasks requiring abstarct and visual reasoning.  It encourages more exploration for future research in finding innovative ideas and frameworks to enhance the reasoning capabilities of future LLMs.

\newpage
\bibliography{anthology,custom}
\bibliographystyle{acl_natbib}

\appendix



\end{document}